\RequirePackage{fix-cm}
\documentclass[twocolumn]{svjour3}          
\smartqed  
\usepackage{graphicx}
\usepackage{latexsym}

\usepackage{url}
\usepackage[T1]{fontenc}
\usepackage{graphicx}
\usepackage{booktabs}
\usepackage{multirow}
\usepackage{xspace}
\usepackage{rotating}
\usepackage[colorinlistoftodos]{todonotes}
\usepackage{soul} 
\usepackage{comment}
\usepackage{adjustbox}
\usepackage{hyperref}
\usepackage[caption=false]{subfig}

\newcommand{\multisubs}{\texttt{MultiSubs}\xspace}

\makeatletter
\DeclareRobustCommand\onedot{\futurelet\@let@token\@onedot}
\def\@onedot{\ifx\@let@token.\else.\null\fi\xspace}

\newcommand{\eg}{\emph{e.g}\onedot} 
\newcommand{\ie}{\emph{i.e}\onedot}

\newcommand{\etal}{\emph{et al}\onedot}

\graphicspath{{figures/}}

\begin{document}

\title{\multisubs: A Large-scale Multimodal and Multilingual Dataset
}
%



\author{Josiah Wang \and
        Pranava Madhyastha \and 
        Josiel Figueiredo \and
        Chiraag Lala \and
        Lucia Specia
}


\institute{J. Wang, P. Madhyastha, L. Specia, C. Lala \at
             Imperial College London, London, UK
           \and
             J. Figueiredo \at 
             Federal University of Mato Grosso, Cuiab\'{a}, Brazil
}

\date{}

\maketitle

\begin{abstract}
This paper introduces a large-scale multimodal and multilingual dataset that aims to facilitate research on grounding words to images in their contextual usage in language. The dataset consists of images selected to unambiguously illustrate concepts expressed in sentences from movie subtitles. The dataset is a valuable resource as (i) the images are aligned to text fragments rather than whole sentences; (ii) multiple images are possible for a text fragment and a sentence; (iii) the sentences are free-form and real-world like; (iv) the parallel texts are multilingual. We set up a fill-in-the-blank game for humans to evaluate the quality of the automatic image selection process of our dataset. We show the utility of the dataset on two automatic tasks: (i) fill-in-the blank; (ii) lexical translation. Results of the human evaluation and automatic models demonstrate that images can be a useful complement to the textual context. The dataset will benefit research on visual grounding of words especially in the context of free-form sentences, and can be obtained from \url{https://doi.org/10.5281/zenodo.5034604} under a Creative Commons licence.\\ 
\end{abstract}

\section{Introduction}
\label{sec:introduction}

``Our experience of the world is multimodal -- we see objects, hear sounds, feel texture, smell odours, and taste flavours'' \cite{BaltrusaitisEtAl:2019}. In order to understand the world around us, we need to be able to interpret such multimodal signals together. Learning and understanding languages is not an exception: humans make use of multiple modalities when doing so. In particular, words are generally learned with visual (among others) input as additional modality. Research on computational models of language grounding using visual information has led to many interesting applications, such as Image Captioning \cite{VinyalsEtAl:2015}, Visual Question Answering \cite{AntolEtAl:2015} and Visual Dialog \cite{DasEtAl:2017}.

Various multimodal datasets comprising images and text have been constructed for different applications. Many of these are made up of images annotated with text labels, and thus do not provide a context in which to apply the text and/or images. More recent datasets for image captioning~\cite{ChenEtAl:2015,YoungEtAl:2014} go beyond textual labels and annotate images with sentence-level text. While these sentences provide a stronger context for the image, they suffer from one primary shortcoming: Each sentence `explains' an image given as a whole, while most often focusing on only some of the elements depicted in the image. This makes it difficult to learn correspondences between elements in the text and their visual representation. Indeed, the connection between images and text is multifaceted, \ie the former is not strictly a textual representation of the latter, thus making it hard to describe a whole image in a single sentence or to illustrate a whole sentence with a single image. A tighter, local correspondence between images and text segments is therefore needed in order to learn better groundings between words and images. Additionally, the texts are limited to very specific domains (image descriptions), while the images are also constrained to very few and very specific object categories or human activities; this makes it very hard to generalise to the diversity of possible real-world scenarios.

\begin{figure}[t]
    \centering
    \includegraphics[width=1.0\linewidth]{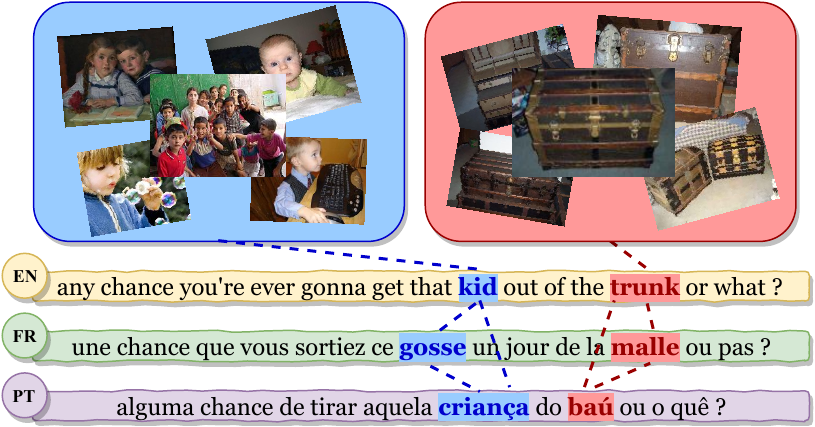}
    \caption{An example instance from our proposed large-scale multimodal and multilingual dataset. \multisubs comprises predominantly conversational or narrative texts from movie subtitles, with text fragments illustrated with images and aligned across languages. 
    }
    \label{fig:overview}
\end{figure}

\begin{figure*}[t]
    \centering
    \includegraphics[width=0.9\linewidth]{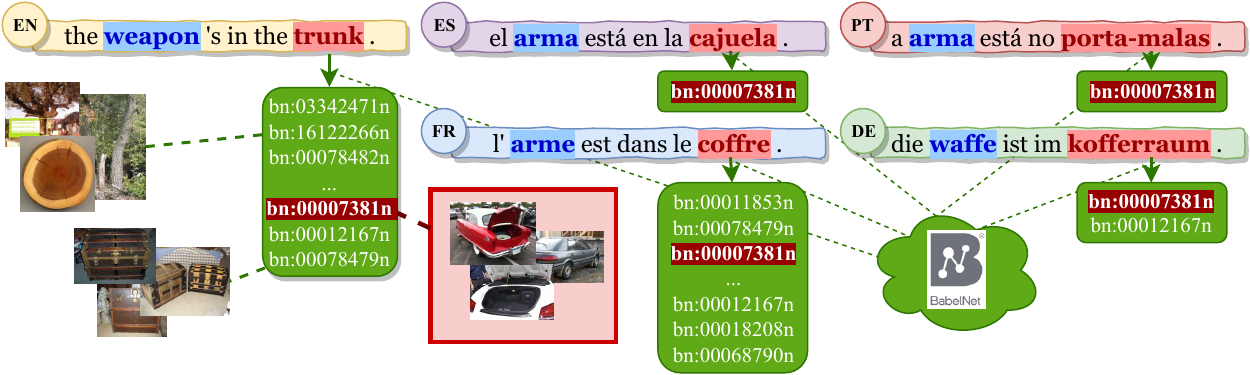}
    \caption{Overview of the \multisubs construction process. Starting from parallel corpora, we selected `visually salient' English words  (\textit{weapon} and \textit{trunk} in this example). We automatically align the words across languages (e.g.\ \textit{trunk} to \textit{cajuela}, \textit{coffre} etc.), and queried BabelNet with the words to obtain a list of synsets. In this example, \textit{trunk} in English is ambiguous, but \textit{cajuela} in Spanish is not. We thus disambiguated the sense of \textit{trunk} by finding the intersection of synsets across languages (bn:00007381n), and illustrate \textit{trunk} with images associated with the intersecting synset, as provided by BabelNet.}
    \label{fig:framework}
\end{figure*}

In this paper we propose \multisubs, a new \textbf{large-scale multimodal and multilingual dataset} that facilitates research on grounding words to images in the \emph{context} of their corresponding sentences (Figure~\ref{fig:overview}). In contrast to previous datasets, ours ground words not only to images but also to their contextual usage in language, potentially giving rise to deeper insights into real-world human language learning. More specifically, (i) text fragments and images in \multisubs have a tighter \emph{local} correspondence, facilitating the learning of associations between text fragments and their corresponding visual representations; (ii) the images are more general and diverse in scope and not constrained to particular domains, in contrast to image captioning datasets; (iii) multiple images are possible for each given text fragment and sentence; (iv) the text comprises a grammar or syntax similar to free-form, real-world text; and (v) the texts are multilingual and not just monolingual or bilingual. Starting from a parellel corpus of movie subtitles ($\S$\ref{sec:subtitles}), we propose a \textbf{crosslingual multimodal disambiguation method} to illustrate text fragments by exploiting the parallel multilingual texts 
to disambiguate the meanings of words in the text (Figure~\ref{fig:framework})  ($\S$\ref{sec:phrase-illustration}). To the best of our knowledge, this has not been previously explored in the context of text illustration. 
We also evaluate the quality of the dataset and illustrated text fragments via human judgment by casting it as a game ($\S$\ref{sec:gapfilling-human}).

We propose and demonstrate two different multimodal applications using \multisubs:
\begin{enumerate}
\item A \textbf{fill-in-the-blank} task to guess a missing word from a sentence, with or without image(s) of the word as clues ($\S$\ref{sec:gapfilling}). 
\item \textbf{Lexical translation}, where we translate a source word in the context of a sentence to a target word in a foreign language, given the source sentence and zero or more images associated with the source word ($\S$\ref{sec:mlt}). 
\end{enumerate}

The dataset can be obtained from \url{https://doi.org/10.5281/zenodo.5034604} under a Creative Commons licence.

\section{Related work}
\label{sec:related}


Most existing multimodal grounding datasets consist of images/videos annotated with noun labels\footnote{Other modalities include speech, audio, etc., but we focus our discussion only on images and text in this paper.} \cite{DengEtAl:2009,LinEtAl:2014}. The main applications of these datasets include multimedia annotation/indexing/retrieval~\cite{SnoekWorring:2005} and object recognition/detection~\cite{LinEtAl:2014,RussakovskyEtAl:2015}. They also enable research on grounded semantic representation or concept learning~\cite{Baroni:2016,BeinbornEtAl:2018}. Besides nouns, other work and datasets focus on labelling and recognising actions~\cite{GellaKeller:2017} and verbs~\cite{GellaEtAl:2016}. These works
, however, are limited to single word labels independent of a contextual usage. 

Recently multimodal grounding work has been moving beyond textual labels to include 
free-form sentences or paragraphs. Various datasets were constructed for these tasks, including image and video descriptions~\cite{BernardiEtAl:2016,AafaqEtAl:2018}, news articles \cite{FengLapata:2010,HollinkEtAl:2016,RamisaEtAl:2018}, 
cooking recipes \cite{MarinEtAl:2018}, 
among others. These datasets, however, ground whole images to the whole text, and making it difficult to identify correspondences between text fragments and elements in the image. In addition, the text also does not explain all elements in the images. 

Apart from monolingual text, there has also been work on multimodal grounding on multilingual text. One primary application of such work is in bilingual lexicon induction using visual data~\cite{KielaEtAl:2015}, where the task is to find words in different languages sharing the same meaning. Hewitt~\etal~\cite{HewittEtAl:2018} has recently developed a large-scale dataset to investigate bilingual lexicon learning for 100 languages. However, this dataset is limited to single word tokens; no textual context is provided with the words. Beyond word tokens, there are also multilingual datasets that are provided at sentence level, primarily extended from existing image description/captioning datasets~\cite{ElliottEtAl:2016,MiyazakiShimizu:2016}. 
Schamoni~\etal~\cite{SchamoniEtAl:2018} also introduce a dataset with images from Wikipedia and their captions in multiple languages; however, the captions are not parallel across languages. These datasets are either very small or use machine translation to generate texts in a different language. More important, 
they are literal descriptions of images gathered for a specific set of object categories or activities and written by users in a constrained setting (\textit{A woman is standing beside a bicycle with a dog}). Like monolingual image descriptions, whole sentences are associated with whole images. This makes it hard to ground image elements to text fragments. 

\section{Corpus and text fragment selection}
\label{sec:subtitles}

\multisubs is based on the OpenSubtitles 2016 (OPUS) corpus~\cite{LisonTiedemann:2016}, which is a large-scale dataset of movie subtitles in 65 languages obtained from OpenSubtitles~\cite{Opensubtitles:2019}. We use a subset by restricting the movies\footnote{We use the term `movie' to cover all types of shows such as movies, TV series, and mini series.} 
 to five categories that we believe are potentially more `visual': adventure, animation, comedy, documentaries, and family. The mapping of IMDb identifiers (used in OPUS) to their corresponding categories are obtained from IMDb's official list~\cite{IMDb:2019}. Most of the subtitles are conversational (dialogues) or narrative (story narration or documentaries). 

The subtitles are further filtered 
to only a subset of English subtitles that has been aligned in OPUS to subtitles from at least one of the top 30 non-English languages in the corpus. This resulted in 45,482 movie instances overall with $\approx$38M English sentences. The number of movies ranges from 2,354 to 31,168 for the top 30 languages.


We aim to select text fragments that are potentially `visually depictable', and which can therefore be illustrated with images. We start by chunking the English subtitles\footnote{PoS tagger from spaCy v2: \texttt{en\_core\_web\_md} from \url{https://spacy.io/models/en}.} to extract nouns, verbs, compound nouns, and simple adjectivial noun phrases. The fragments are ranked by imageability scores obtained via bootstrapping from the MRC Psycholinguistic database~\cite{PaetzoldSpecia:2016}; for multi-word phrases we average the imageability score of each individual word, assigning a zero score to each unseen word. We retain 
text fragments with an imageability score of at least 500, which is determined by manual inspection of a subset of words. After removing fragments occurring only once, the output is a set of 144,168 unique candidate fragments (more than 16M instances) 
across $\approx$11M sentences. 

\section{Illustration of text fragments}
\label{sec:phrase-illustration}



Our 
approach for illustrating \multisubs obtains images for a subset of text fragments: 
\emph{single word nouns}.
Such nouns occur substantially more often in the corpus and are thus more suitable for learning algorithms. 
Additionally, single nouns (\emph{dog}) make it more feasible to obtain good representative images than longer phrases (\emph{a medium-sized white and brown dog}). This filtering step results in 4,099 unique English nouns occurring in $\approx$10.2M English sentences.

We aim to obtain images that illustrate the correct \emph{sense} of these nouns in the context of the sentence. For that, we propose a novel approach that exploits the aligned multilingual subtitle corpus 
for sense disambiguation using BabelNet~\cite{NavigliPonzetto:2012} ($\S$\ref{sec:phrase-disambiguation}), a multilingual sense dictionary. 
 Figure~\ref{fig:framework} illustrates the process.
 
 \multisubs is designed as a subtitle corpus illustrated with \emph{general} images. Taking images from the video from where the subtitle comes is not possible since we do not have access to the copyrighted materials. In addition, there are no guarantees that the concepts mentioned in the text would be depicted in the video.


\subsection{Cross-lingual sense disambiguation}
\label{sec:phrase-disambiguation}

The key intuition to our proposed text illustration approach is that an ambiguous English word may be unambiguous in the parallel sentence in the target language. For example, the correct word sense of \textit{drill} in an English sentence can be inferred from a parallel Portuguese sentence based on the occurrence of the word \textit{broca} (the machine) or \textit{treino} (training exercise).

\paragraph{Cross-lingual word alignment.} We experiment with up to four \emph{target} languages in selecting the correct images to illustrate our candidate text fragments (nouns): \textbf{Spanish} (\textbf{ES}) and \textbf{Brazilian Portuguese} (\textbf{PT}), which are the two most frequent languages in OPUS; and \textbf{French} (\textbf{FR}) and \textbf{German} (\textbf{DE}), both commonly used in existing Machine Translation (MT) and Multimodal Machine Translation (MMT) research~\cite{ElliottEtAl:2017}. For each 
language, subtitles are selected such that (i) each is aligned with a subtitle in English; (ii) each contains at least one noun of interest.

For English and each target language, we trained \texttt{fast\_align}~\cite{DyerEtAl:2013} on the \emph{full} set of parallel sentences (regardless of whether the sentence contains a candidate fragment) to obtain alignments between words in both languages (symmetrised by the intersection of alignments in both directions). This generates a dictionary which maps English nouns to words in the target language. We filter this dictionary to remove pairs with infrequent target phrases (under 1\% of the corpus).
We also group words in the target language that share the same lemma\footnote{We used the lemmas provided by spaCy.}.

\paragraph{Sense disambiguation.} A noun being translated to different words in the target language does not necessarily mean it is ambiguous. The target phrases may simply be synonyms referring to the same concept. Thus, we further attempt to group synonyms on the target side, while also determining the correct word sense by looking at the aligned phrases across \emph{multilingual} corpora. 

For word senses, we use BabelNet~\cite{NavigliPonzetto:2012}, which is a large semantic network and multilingual encyclopaedic dictionary covering many languages and unifies 
other semantic networks. We query BabelNet with the English noun and 
its possible translations in each target language from our automatically aligned dictionary. 
The output (queried separately per language) is a list of BabelNet synset IDs matching the query.  

To help us identify the correct sense of an English noun for a given context, we use the aligned word in the parallel sentence in the target language for disambiguation. We compute the intersection between the BabelNet synset IDs returned from both queries. For example, the English query \emph{bank} could contain the synsets \emph{financial-bank} and \emph{river-bank}, and the Spanish query for the corresponding translation \emph{banco} only returns the synset \emph{financial-bank}. In this case, the intersection of both synset sets allows us to decide that \emph{bank}, when translated to \emph{banco}, refers to its \emph{financial-bank} sense. Therefore, we can annotate the respective parallel sentence in the corpus with the correct sense. Where multiple synset IDs intersect, we take the union of all intersecting synsets as possible senses for the particular alignment. This potentially means that (i) the term is ambiguous and the ambiguity is carried over to the target language; or (ii) the distinct BabelNet synsets actually refer to the same or similar sense, as BabelNet unifies word senses from multiple sources automatically. We name this dataset 
\textbf{\textit{intersect$_1$}}.

If the above is only performed for one language pair, this single target language may not be sufficient to disambiguate the sense of the English term, as the term might be ambiguous in both languages (e.g.\ \textit{coffre} is also ambiguous in Figure~\ref{fig:framework}). This is particularly true for closely related languages such as Portuguese and Spanish. Thus, we propose exploiting \emph{multiple} target languages to further increase our confidence in disambiguating the sense of the English word. Our assumption is that more languages will eventually allow the correct context of the word to be identified. 



More specifically, we examine subtitles containing parallel sentences for up to four target languages. For each English phrase, we retain instances with at least one intersection between the synset IDs across all $N$ languages, and discard 
if there is no intersection. We name these datasets \textbf{\textit{intersect$_N$}}, which comprise sentences that have valid synset alignments to at least $N$ languages. Note that \textbf{\textit{intersect$_{N+1}$}} $\subseteq$ \textbf{\textit{intersect$_N$}}.

Table~\ref{tbl:stats-dataset-size} shows the final dataset sizes of \textbf{\textit{intersect$_N$}}. 
Our experiments in $\S$\ref{sec:experiments} will focus on comparing models trained on different \textbf{\textit{intersect$_N$}} datasets, and test them on \textbf{\textit{intersect$_4$}}. 


\begin{table}[t]
\caption{Number of sentences for the \textbf{\textit{intersect$_N$}} subset of \multisubs,    where $N$ is the minimum number of target languages used for disambiguation. The slight variation in the final column is due to differences in how the aligned sentences are combined or split in OPUS across languages.}
	\label{tbl:stats-dataset-size}
\small
	\centering
		\begin{tabular}{@{}ccccc@{}}
			\toprule
              \noalign{\smallskip}
			  & $N=1$ & $N=2$ & $N=3$ & $N=4$  \\
			\midrule
 ES & 2,159,635 & 1,083,748 & 335,484 & 45,209 \\
 PT & 1,796,095 & 1,043,991 & 332,996 & 45,203 \\
 FR & 1,063,071 & 641,865 & 305,817 & 45,217 \\
 DE & 384,480 & 250,686 & 131,349 & 45,214 \\
			\bottomrule
		\end{tabular}
\end{table}

\paragraph{Image selection.} The final step to constructing \multisubs is to assign at least one image to each disambiguated English term, and by design the term in the aligned target language(s). As BabelNet generally provides multiple images for a given synset ID, 
we illustrate the term with all Creative Commons images associated with the synset.

\section{MultiSubs statistics and analysis}
\label{sec:statistics}

Table~\ref{tbl:stats-dataset-size} shows the number of sentences in \multisubs, according to their degree of 
intersection. On average, there are 1.10 illustrated words per sentence in 
\multisubs, where about 90-93\% sentences contain one illustrated word per sentence (depending on the target language). The number of images for each BabelNet synset ranges from 1 to 259, with an average of 15.5 images (excluding those with no images). 

Table~\ref{tbl:tokentype} shows some statistics of the sentences in \multisubs. \multisubs is substantially larger and less repetitive than Multi30k~\cite{ElliottEtAl:2016} ($\approx$300k tokens, $\approx$11-19k types, and only $\approx$5-11k singletons), even though the sentence length remains similar.


\begin{table}
\caption{Token/type statistics on the sentences of \textbf{\textit{intersect$_1$}} \multisubs.}
\label{tbl:tokentype}
\small
\centering
\begin{tabular}{lcccc}
\toprule
& \textbf{tokens} & \textbf{types} & \textbf{avg length} & {\bf singletons} \\
\midrule
EN & 27,423,227 & 152,520 & 12.70 & 2,005,874\\
ES & 25,616,482 & 245,686 & 11.86 & 2,012,476\\
\midrule
EN & 23,110,285 & 138,487 & 12.87 & 1,685,102\\
PT & 20,538,013 & 205,410 & 11.43 & 1,687,903\\
\midrule
EN & 13,523,651 & 104,851 & 12.72 & 1,012,136\\
FR & 12,956,305 & 149,372 & 12.19 & 1,004,304\\
\midrule
EN & 4,670,577 & 62,138 & 12.15 & 364,656\\
DE & 4,311,350 & 123,087 & 11.21 & 364,613\\
\bottomrule
\end{tabular}
\end{table}

\begin{figure}
\centering
\resizebox{\linewidth}{!}{ 
\begin{tabular}{@{}lr@{}}
& \multirow{6}{*}{\includegraphics[width=0.23\linewidth]{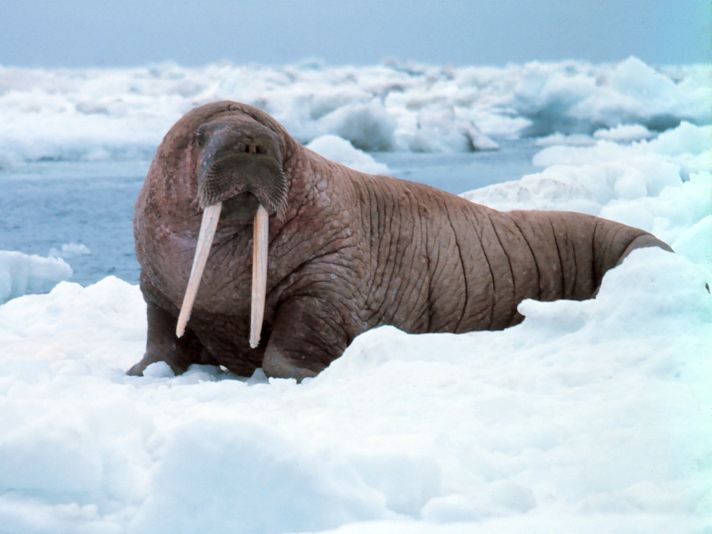}} \\
\textit{bn:00070012n (seal wax), bn:00070013n (stamp),} & \\
\textit{bn:00070014n (sealskin), ...} & \\
\textit{EN}: stamp my heart with a \textbf{seal} of love ! & \\
\textit{FR}: frapper mon cœur d' un \textbf{sceau} d' amour ! & \\ 
& \\
\midrule
& \multirow{8}{*}{\includegraphics[width=0.20\linewidth]{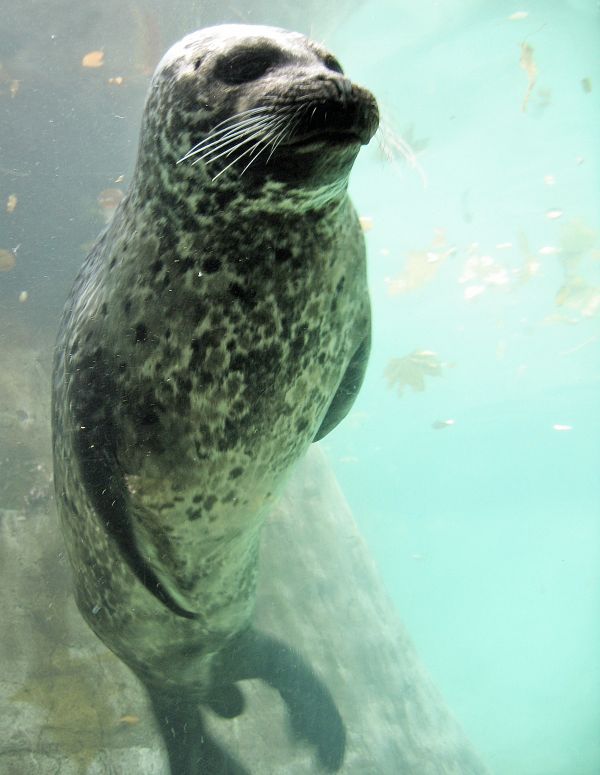}} \\
\textit{bn:00021163n (animal)} & \\
\textit{EN}: even the \textbf{seal} 's got the badge . & \\
\textit{ES}: que hasta la \textbf{foca} tiene placa . & \\
\textit{PT}: até a \textbf{foca} tem um distintivo . &\\
\textit{FR}: même l' \textbf{otarie} a un badge . &\\
\textit{DE}: sogar die \textbf{robbe} hat das abzeichen . & \\
\end{tabular}
}
\caption{Example of using multilingual corpora to disambiguate and illustrate a phrase.}
\label{fig:examplecleaning}
\end{figure}

\begin{figure}[t]
    \centering
    \resizebox{\linewidth}{!}{ 
    \begin{tabular}{@{}l r@{}}
        they knew the gods put dewdrops on \underline{plants} &  \multirow{4}{*}{\includegraphics[height=1.5cm]{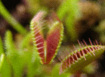}}\\
        in the night. & \\
        sabiam que os deues punham orvalho nas & \\
        \underline{plantas} \`{a} noite & \\
        \noalign{\bigskip}
        today we are announcing the closing of 11 of &  \multirow{4}{*}{\includegraphics[height=1.5cm]{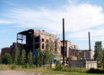}}\\
        our older \underline{plants}. & \\
        hoje anunciamos o encerramento de 11 das & \\
         \underline{f\'{a}bricas} mais antigas. & \\
    \end{tabular}
    }
    \caption{Example disambiguation in the EN-PT portion of \multisubs. In both cases, \textit{plants} were correctly disambiguated using 4 languages.}
    \label{fig:egdisambiguation}
\end{figure}

Figure~\ref{fig:examplecleaning} shows an example of how multilingual corpora is beneficial for disambiguating the correct sense of a word and subsequently illustrating it with an image. The top example 
shows an instance from 
\textbf{\emph{intersect$_1$}},
where the English sentence is aligned to only one target language (French). In this case, the word \textbf{sceau} is ambiguous in BabelNet, covering different but mostly related senses, and in some cases is noisy (terms are obtained by automatic translation). 
The bottom example shows an example where the English sentence is aligned to four target languages, which came to a consensus on a single BabelNet synset (and illustrated with the correct image).
A manual inspection of a randomly selected subset of the data to assess 
our automated disambiguation procedure showed that \textbf{\emph{intersect$_4$}} is of high quality. We found many interesting cases of ambiguities, some of which are shown in Figure~\ref{fig:egdisambiguation}. 




\section{Human evaluation}
\label{sec:gapfilling-human}

To quantitatively assess our automated cross-lingual sense disambiguation cleaning procedure
, we collect human annotations to determine whether images in \multisubs are indeed useful for predicting a missing word in a fill-in-the-blank task. The annotation also serves as a human upperbound to the task (detailed in $\S$\ref{sec:gapfilling}), measuring whether images are useful for helping humans guess the missing word.

\begin{figure}
    \centering
    \includegraphics[width=0.95\linewidth]{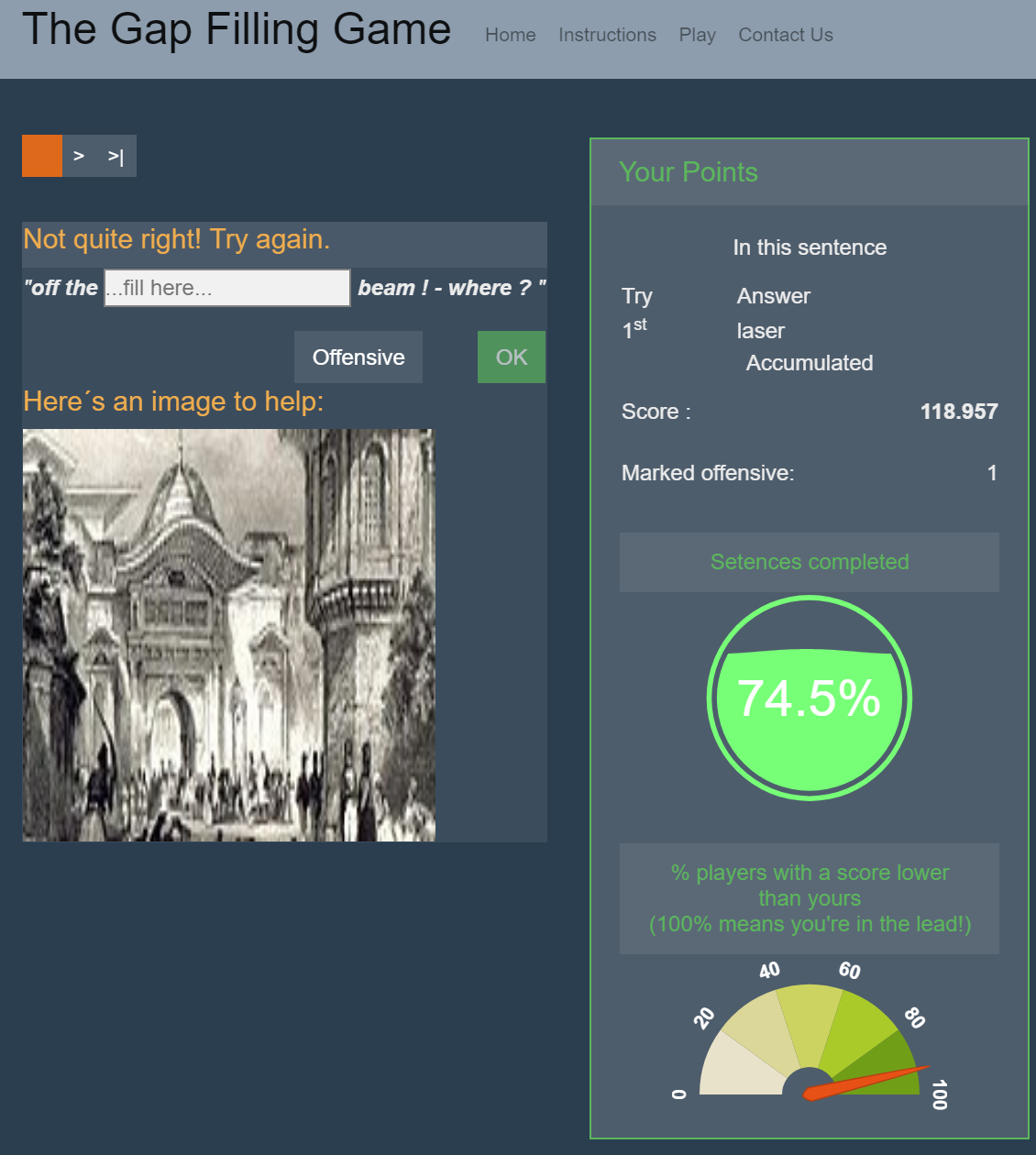}
    \caption{A screenshot of \textit{The Gap Filling Game}, used to evaluate our automated cleaning procedure, as an upperbound to how well humans can perform the task without images, and to evaluate whether images are actually useful for the task.}
    \label{fig:interface}
\end{figure}

We set up the annotation task as \textit{The Gap Filling Game} (Figure~\ref{fig:interface}). In this game, users are given three attempts at guessing the exact word removed from a sentence from \multisubs. In the first attempt, the game shows only the sentence (along with a blank space for the missing word). In the second attempt, the game additionally provides one image for the missing word as a clue. In the third and final attempt, the system shows all images associated with the missing word. At each attempt, users are awarded a score of 1.0 if the word they entered is an exact match to the original word, or otherwise a partial score (between 0.0 and 1.0) computed as the cosine similarity between pre-trained CBOW word2vec~\cite{MikolovEtAl:2013} embeddings of the predicted and the original word. 
Each `turn' (one sentence) ends when the user enters an exact match or after he or she has exhausted all three attempts, whichever occurs first. The score at the second and third attempts are multiplied by a \emph{penalty factor} (0.90 and 0.80 respectively) to encourage users to guess the word correctly as early as possible. A user's score for a single turn is the maximum over all three attempts, and the final cumulative score per user is the sum of the score across all annotated sentences. This final score determines the winner and runner-up at the end of the game (after a pre-defined cut-off date), both of whom are awarded an Amazon voucher each. Users are not given an exact `current top score' table during the game, but are instead provided the percentage of all users who has a lower score than the user's current score.

For the human annotations, we also introduce the \textbf{$intersect_0$} dataset where the words are not disambiguated, i.e.\ images from all matching BabelNet synsets are used. This is to evaluate the quality of our automated filtering process. Annotators are allocated 100 sentences per batch, and are able to request for more batches once they complete their allocated batch. Sentences are selected at random. 
To select one image for the second attempt, we select the image most similar to the majority of other images of the synset, by computing the cosine distance of each image's ResNet152 pool5 feature~\cite{HeEtAl:2016} 
against all remaining images in the synset, and averaged the distance across these images. 

Users are allowed to complete as many sentences as they like. The annotations were collected over 24 days in December 2018, and participants are primarily staff and student volunteers from the University of Sheffield, UK. 
238 users participated in our annotation, resulting in 11,127 annotated instances (after filtering out invalid annotations).

\paragraph{Results of human evaluation}

Table~\ref{tbl:stats_humanevalexact} shows the results of human annotation, comparing the proportion of instances correctly predicted by annotators at different attempts: (1) no image; (2) one image; (3) many images; and also those that fail to be correctly predicted after three attempts. We consider a prediction correct if the predicted word is an exact match to the original word. Overall, out of $11,127$ instances, $21.89\%$ of instances were predicted correctly with only the sentence as context, $20.49\%$ with one image, and $15.21\%$ with many images. The annotators failed to guess the remaining $42.41\%$ of instances. Thus, we can estimate a human upper bound of $57.59\%$ for correctly predicting missing words in the dataset, regardless of the cue provided. Across different $intersect_N$ splits, there is an improvement in the proportion of correct predictions as $N$ increases, from $54.55\%$ for $intersect_0$ to $60.83\%$ for $intersect_4$. We have also tested sampling each split to have an equal number of instances ($1,598$) to ensure that the proportion is not an indirect effect of imbalance in the number of instances; we found the proportions to be similar.

\begin{table*}[t]
    \caption{Distribution across different attempts by humans in the fill-in-the-blank task.}
    \label{tbl:stats_humanevalexact}
    \centering
    \small
    \begin{tabular}{cccccc}
    \toprule
    & \multicolumn{4}{c}{Correct at attempt} &
    \multirow{2}{*}{Total} \\
    \noalign{\smallskip}    
    \cline{2-5}
    \noalign{\smallskip}
    & 1 & 2 & 3 & Failed & \\
    \midrule
    $intersect_0$ & 611 (18.75\%) & 660 (20.26\%) & 503 (15.44\%) & 1484 (45.55\%) & 3258\\
    $intersect_1$ & 534 (21.86\%) & 481 (19.69\%) & 378 (15.47\%) & 1050 (42.98\%) & 2443\\
    $intersect_2$ & 462 (22.35\%) & 408 (19.74\%) & 303 (14.66\%) & 894 (43.25\%) & 2067\\
    $intersect_3$ & 432 (24.53\%) & 388 (22.03\%) & 260 (14.76\%) & 681 (38.67\%) & 1761\\
    $intersect_4$ & 397 (24.84\%) & 343 (21.46\%) & 248 (15.52\%) & 610 (38.17\%) & 1598\\
    \midrule
    all & 2436 (21.89\%) & 2280 (20.49\%) & 1692 (15.21\%) & 4719 (42.41\%) & 11127\\
    \bottomrule
    \end{tabular}
\end{table*}

A user might fail to predict the exact word, but the word might be semantically similar to the correct word (e.g.\ a synonym). Thus, we also evaluate the annotations with the cosine similarity between word2vec embeddings of the predicted and correct word. Table~\ref{tbl:stats_humanevalscore} shows the average word similarity scores at different attempts across 
$intersect_N$ splits. Across attempts, 
the average similarity score is lowest for attempt 1 (text-only, 0.36), compared to attempts 2 (one image) and 3 (many images) -- 0.48 and 0.49 respectively. Again, we verified that the scores are not affected by the imbalanced number of instances, by sampling equal number of instances across splits and attempts. We also observe a generally higher average score as we increase $N$, albeit marginal.

\begin{table}[t]
    \centering
    \caption{Average word similarity scores of human evaluation of the fill-in-the-blank task.}
    \label{tbl:stats_humanevalscore}
    \begin{tabular}{cccc}
        \toprule
        & \multicolumn{3}{c}{Average scores for attempt} \\
        \cline{2-4}
        & 1 & 2 & 3 \\
        \midrule
        $intersect_0$ & 0.33 (3258) & 0.47 (2647) & 0.47 (1987) \\
        $intersect_1$ & 0.36 (2443) & 0.47 (1909) & 0.49 (1428) \\
        $intersect_2$ & 0.37 (2067) & 0.48 (1605) & 0.48 (1197) \\
        $intersect_3$ & 0.38 (1761) & 0.51 (1329) & 0.50 (941) \\
        $intersect_4$ & 0.39 (1598) & 0.50 (1201) & 0.52 (858) \\
        \midrule
        all & 0.36 (11127) & 0.48 (8691) & 0.49 (6411) \\
        \bottomrule
    \end{tabular}
\end{table}

\begin{figure}[t]
    \centering
    \begin{tabular}{l}
             he was one of the best pitchers in  \textbf{baseball} .\\
        \textit{baseball} (1.00)\\
         \midrule
         uh , you know , i got to fix the \textbf{sink} , catch the game .\\
         \textit{car} (0.06), \textit{sink} (1.00) \\
         \includegraphics[height=1.2cm]{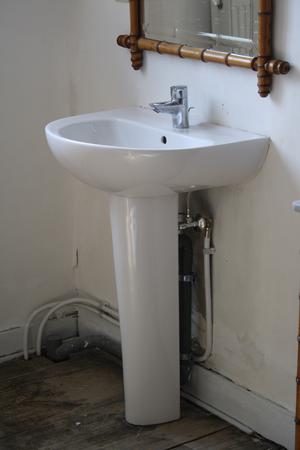} \\        
         \midrule
         i saw it at the \textbf{supermarket} and i thought that maybe \\you would have liked it .\\
         \textit{market} (0.18) \textit{shop} (0.50), \textit{supermarket} (1.00)\\   
         \includegraphics[height=1.2cm]{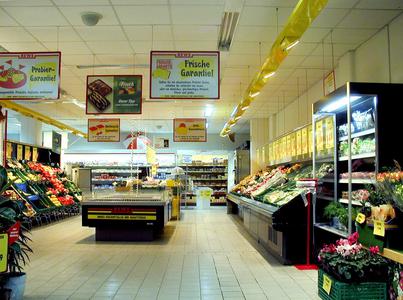}     
         \includegraphics[height=1.2cm]{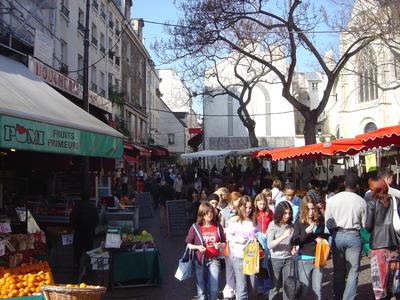}             
         \includegraphics[height=1.2cm]{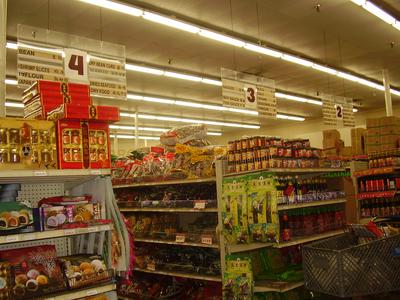} \includegraphics[height=1.2cm]{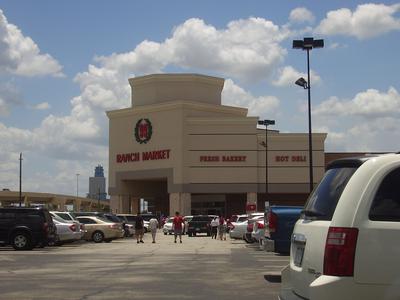} \\ 
         \midrule
         It's mac , the night \textbf{watchman} .\\
         \textit{before} (0.07), \textit{police} (0.31), \textit{guard} (0.26)\\
         \includegraphics[height=1.2cm]{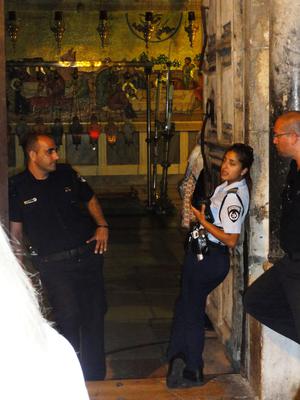}
         \includegraphics[height=1.2cm]{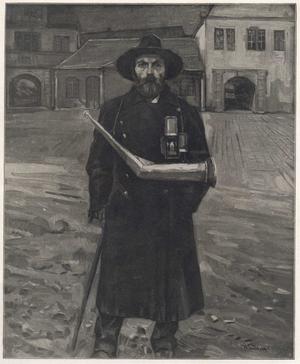}
         \includegraphics[height=1.2cm]{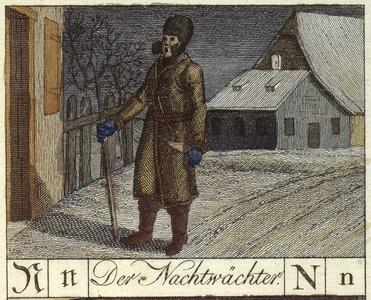}         
         \includegraphics[height=1.2cm]{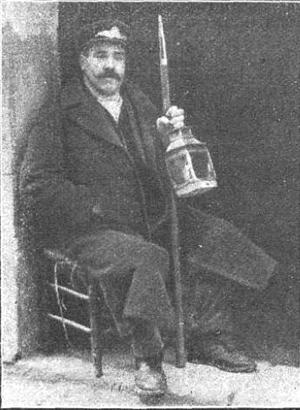}
         \includegraphics[height=1.2cm]{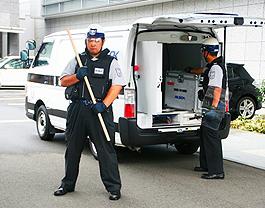}         
    \end{tabular}
    \caption{Example annotations from our human experiment, with the masked word \textbf{boldfaced}. Users' guesses are \textit{italicised}, with the word similarity score in brackets. The first example was guessed correctly without any images. The second was guessed correctly after one image was shown. The third  was only guessed correctly after all images were shown. The final example was not guessed correctly after all three attempts.}
    \label{fig:human-example}
\end{figure}

Figure~\ref{fig:human-example} shows a few example human annotations, with varying degrees of success. In some cases, textual context alone is sufficient for predicting the correct word. In other cases, like in the second example, it is difficult to guess the missing word purely from textual context. In this case, images are useful.

We conclude that the task of filling in the blanks in \multisubs is quite challenging even for humans, where only $57.59\%$ instances were correctly guessed. This inspired us to introduce fill-in-the-blank as a task to evaluate how well automatic models can perform the same task, with or without images as cues ($\S$\ref{sec:gapfilling}).

\section{Experimental evaluation}
\label{sec:experiments}

We demonstrate how \multisubs can be used to train models to learn multimodal text grounding with images on two different applications. 

\subsection{Fill-in-the-blank task}
\label{sec:gapfilling}

The first task we present for \multisubs is a fill-in-the-blank task. 
The objective of the task is to predict a word that has been removed from a sentence in \multisubs, given the masked sentence as \emph{textual context} and optionally one or more images depicting the missing word as \emph{visual context}. Our hypothesis is that images in \multisubs can provide additional contextual information complementary to the masked sentence, and that models that utilize both textual and visual contextual cues will be able to recover the missing word better than models that use either alone. Formally, given a sequence $S$$=$$\{w_1, \ldots, w_{t-1}, w_{t}, w_{t+1}, \ldots, ... w_{T}\}$ of length $T$, where $w_{t}$ is unobserved while the others are observed, the task is to predict $w_{t}$ given $S$ and optionally one or more images $\{I_{1}, I_{2}, \ldots, I_{K}\}$. 

This task is similar to the human annotation ($\S$\ref{sec:gapfilling-human}). Thus, we use the statistics from human evaluation as an estimated human upperbound for the task. We observe that this task is challenging even for humans who successfully predicted the missing word for only $57.59\%$ of instances, regardless of whether they use images as contextual cue.

\subsubsection{Models}
\label{sec:gapfilling-models}

We train three computational models for this task: 
(i) a \textbf{text-only} model, where we follow Lala~\etal~\cite{LalaEtAl:2019} and use a bidirectional recurrent neural network (BRNN) that takes in the sentence with blanks and predicts at each time step either \texttt{null} or the word corresponding to the blank;  
%
%
(ii) an \textbf{image-only} baseline, where we treat the task as image labelling and build a two layer feed-forward neural network, with ResNet152 pool5 layer~\cite{HeEtAl:2016} as image features; and 
%
(iii) a \textbf{multimodal} model, where we follow Lala~\etal~\cite{LalaEtAl:2019} and use simple multimodal fusion to initialize the BRNN model with the image features.   


\subsubsection{Dataset and settings}
\label{sec:gapfilling-settings}

We blank out each 
illustrated word of a sentence as a fill-in-the-blank instance. If a sentence contains multiple illustrated nouns, we replicate the sentence and generate a blank per sentence for each noun, treating each as a separate instance. 

The number of validation and test instances is fixed at $5,000$ each. These comprise sentences from \textbf{$intersect_4$}, which we consider to be the cleanest subset of \multisubs. The validation and test sets are made more challenging by (i) uniformly sampling nouns from \textbf{$intersect_4$} to increase their diversity; (ii) sampling an instance for each possible BabelNet sense of a sampled noun; this increases the semantic (and visual) variety for each word (\eg sampling both the financial institution sense and the river sense of the noun `bank'). The training set comprises all remaining instances. 

We sample one image at random from the corresponding synset to illustrate each sense-disambiguated noun. Our preliminary analysis showed that, in most cases, an image tends to correspond to only a single word label. This makes it less challenging for an image classifier which simply performs an exact matching of a test image to a training image, as the same image is repeated frequently across instances of the same noun. To circumvent this problem, we ensured that the images in the validation and test sets are both disjoint from the images in the training set. This is done by reserving 10\% of all unique images for each synset in the validation and test sets respectively, and removing all these images from the training set. Our final training set consists of $4,277,772$ instances with $2,797$ unique masked words. The number of unique words in the validation and test set is $496$ and $493$ respectively, signifying their diversity.


\subsubsection{Evaluation metrics}
\label{sec:gapfilling-metrics}

The models are evaluated using two metrics: (i) accuracy; (ii) average word similarity. The \textbf{accuracy} measures the proportion of correctly predicted words (exact token match) across test instances. The \textbf{word similarity} score measures the average semantic similarity across test instances between the predicted word and the correct word. For this paper, the cosine similarity between word2vec embeddings is used. 
Our evaluation script can be found on \url{https://github.com/josiahwang/multisubs-eval}

\subsubsection{Results}
\label{sec:gapfilling-results}

We trained different models on four disjoint subsets of the training samples. These are selected such that 
each corresponds to words whose sense have been disambiguated using \textbf{exactly} $N$ languages, i.e.\ $intersect_{=N} \subseteq intersect_{N}$ ($\S$\ref{sec:phrase-disambiguation}). This allows us to investigate whether our automated cross-lingual  disambiguation process helps improve the quality of the images used to unambiguously illustrate the words.

During development, our models encountered issues with predicting words that are unseen in the training set, resulting in lower than expected accuracies. This is especially true when training with the $intersect_{=N}$ subsplits. Thus, we report results on a subset of the full test set that only contains output words that have been seen across all training subsplits $intersect_{=N}$ 
. This test subset contains $3,262$ instances with $169$ unique words, and is used across all training subsets.
\begin{table}[t]
    \caption{Accuracy and word similarity scores for our baseline (text-only) models on the fill-in-the-blank task, evaluated on the test subset and trained on the full training set.}
    \label{tbl:gapfilling-baseline}
    \centering
    \resizebox{0.9\linewidth}{!}{  
    \begin{tabular}{c c c}
    \toprule
    & \textbf{Accuracy (\%)} & \textbf{Word similarity} \\
        \midrule
    random & 0.00 & 0.10\\
    random-multinomial & 0.03 & 0.12\\
    $1$-gram & 1.07 & 0.17\\
    $2$-gram & 8.74 & 0.22\\
    $3$-gram & 16.03 & 0.31\\
    $4$-gram & 23.67 & 0.38\\
    $5$-gram & 27.35 & 0.41\\
    $6$-gram & 29.28 & 0.43\\
    $7$-gram & 30.07 & 0.43\\
    $8$-gram & 30.32 & 0.44\\
    $9$-gram & 30.35 & 0.44\\
    \bottomrule
    \end{tabular}
    }
\end{table}
\paragraph{Baseline models. } 
Our baseline models are (i) a \textbf{random} baseline that predicts a random target word from the full training set; (ii) a \textbf{random-multinomial} baseline that randomly samples a target word based on its frequency distribution in the full training set; (iii) a classic \textbf{$n$-gram} model with back-off. The $n$-gram model learns the most frequent target word from the full training set given the previous $n-1$ context words; the context window is iteratively reduced if the context is not found. In the case of $n=1$, the model always predicts the most frequent blanked-out word (\textit{man} for our dataset). We report results for $n\le9$ (the predictions do not change after $n=9$). 

Table~\ref{tbl:gapfilling-baseline} presents the baseline results on the test subset. As expected, randomly guessing the blank word does not get the system far. It is useful to note that the word similarity score has a lower-bound of 0.10. Always guessing \textit{man} ($1$-gram) is slightly better than randomly guessing, although the accuracy is still low at 1\%. Surprisingly, the simple $n$-gram models with back-off actually perform well, with an accuracy of 23.67\% for $4$-grams and up to 30.35\% for $9$-grams. The word similarity scores show a similar trend, with a maximum score of 0.44 with $9$-grams.

\paragraph{Neural-based models. }
Table~\ref{tbl:gapfilling-accuracy} shows the accuracies of our automatic models on the fill-in-the-blank task, compared to the estimated human upperbound of $57.59\%$. Overall, text-only models perform better than their image-only counterparts. Multimodal models that combine both text and image signals perform better in some cases ($intersect_{=3}$ and $intersect_{=4}$), especially with the cleaner splits where the images have been filtered with more languages. This can be observed when both the performance of image-only models and multimodal models improve as the number of languages used to filter the images increases. This suggests that our automated cross-lingual disambiguation process is beneficial.

The text-only models appear to give lower accuracy as the number of languages increases; this is naturally expected as the size of the training set is smaller for larger number of intersecting languages. However, the opposite is true for our multimodal model -- we observe substantial improvements instead as the number of languages increases (and thus fewer training examples). This demonstrates that the additional image modality actually helped improve the accuracy, even with a smaller training set.

Table~\ref{tbl:gapfilling-similarity} shows the average word similarity scores of the models on the task, to account for predictions that are semantically correct despite not being an exact match. Again, a similar trend is observed: images become more useful as the image filtering process becomes more robust. Thus, we conclude that, given cleaner versions of the dataset, images may prove to be a useful, complementary signal for the fill-in-the-blank task.


\begin{table}[t]
    \caption{Accuracy scores (\%) for the fill-in-the-blank task on the test subset, comparing text-only, image-only, and multimodal model  trained on different subsets of the data. 
    }
    \label{tbl:gapfilling-accuracy}
    \centering
    \begin{tabular}{c c c c}
    \toprule
    & \textbf{text} & \textbf{image} & \textbf{multimodal} \\
        \midrule
          $intersect_{=1}$ & 16.49 & 9.84 & 14.53\\
$intersect_{=2}$ & 18.82 & 11.62 & 16.00\\
$intersect_{=3}$ & 17.72 & 12.91 & 19.19\\       
$intersect_{=4}$ & 15.57 
& 13.70 & 30.53\\
    \bottomrule
    \end{tabular}
\end{table}

\begin{table}[t]
    \caption{Word similarity scores for the fill-in-the-blank task on the test subset, comparing text-only, image-only, and multimodal models trained on different subsets of the data.}
    \label{tbl:gapfilling-similarity}
    \centering
    \begin{tabular}{c c c c}
    \toprule
    & \textbf{text} & \textbf{image} & \textbf{multimodal} \\
        \midrule
 $intersect_{=1}$ & 0.34 & 0.28 & 0.33\\
 $intersect_{=2}$ & 0.36 & 0.29 & 0.34\\
 $intersect_{=3}$ & 0.34 & 0.30 & 0.36\\   
 $intersect_{=4}$ & 0.25 & 0.31 & 0.43\\
    \bottomrule
    \end{tabular}
\end{table}

It is interesting that the complex neural-based models (even the multimodal ones) did not perform better than our simpler $n$-gram based model. This may be because the $n$-gram models are trained on the full dataset rather than the subsplits, although we did not observe any better accuracy when training our BRNN text model on the full training set (17.6\% accuracy). To further investigate this, we compute the statistics for the $n$-gram models on the different $intersection_{=N}$ subsplits. The $9$-gram models still achieved accuracies comparable to the BRNN text model, between 16.1\% to 16.6\% for $intersection_{=1}$, $intersection_{=2}$, and $intersection_{=3}$. The accuracies are already close to this level even for $5$-grams. The $9$-gram model for $intersection_{=4}$ actually achieved a higher accuracy of 22.75\% despite being the smaller subsplit, suggesting that perhaps some of the instances here are useful for predicting the test set; indeed, the test set was taken from this subsplit. There is also a chance that the neural-based models might just need more tweaking to perform better than the $n$-gram models; we leave this as future work to investigate.

\subsection{Lexical translation}
\label{sec:mlt}

As \multisubs is a multimodal \emph{and} multilingual dataset, we explore a second application for \multisubs, that is lexical translation (LT). The objective of the LT task is to translate a given word $w^s$ in a source language~$s$ to a word $w^f$ in a specified target language $f$. The translation is performed in the context of the original sentence in the source language, and optionally with one or more images corresponding to the word $w^s$. 

The prime motivation for this task is to investigate challenges in multimodal machine translation (MMT) at a lexical level, i.e.\ for dealing with words that are ambiguous in the target language, or for tackling out-of-vocabulary words. For example, the word \textit{hat} can be translated to German as \textit{hut} (stereotypical hat with a brim all around the bottom), \textit{kappe} (a cap), or \textit{m{\"u}tze} (winter hat). Textual context alone may not be sufficient to inform translation in such cases, and the hypothesis that images can be used to complement the sentences in helping translate a source word to its correct sense in the target language.

For this paper, we fix English as the source language, and explore translating words to the four target languages in \multisubs: Spanish (ES), Portuguese (PT), French (FR), and German (DE). 

\subsubsection{Models}
\label{sec:mlt-models}
We follow the models as described in $\S$\ref{sec:gapfilling-models}. The text based models are based on a similar BRNN model but in this case, the input contains the entire English source sentence with a marked word (marked with an underscore). The marked word in this case is the word to be translated. The BRNN model is trained with the objective of predicting the translation for the marked word and \texttt{null} for all words that are not marked. Our hypothesis is that in the current setting the model is able to maximally exploit context for the translation of the source word. 

For the image-only model, we use a two-layered neural network where the input is the source word and the image feature is used to condition the hidden layer. 
The multimodal model is a variant of the text-only BRNN, but initialized with the image features.

\subsubsection{Dataset and settings}
\label{sec:mlt-dataset}

We use the same procedure as $\S$\ref{sec:gapfilling-settings} to prepare the dataset. 
Instead of being masked, the English words now act as the word to be translated. The corresponding word in the target language is obtained from our automatic alignment procedure ($\S$\ref{sec:phrase-disambiguation}). For each target language, we use a subset of \multisubs where the sentences are aligned to the target language. We also remove unambiguous instances where a 
word can only be translated to the same word in the target language. 

Like $\S$\ref{sec:gapfilling-settings}, the number of validation and test instances is fixed at $5,000$ each
. Again, to tackle issues with unseen labels, we use a version of the test set which contains only a subset where the output target words are seen during training.

The number of training instances per language are: $2,356,787$ (ES), $1,950,455$ (PT), $1,143,608$ (FR), and $405,759$ (DE). The number of unique source and target words are 
between $131-153$ and $173-223$ respectively for the test set.

Like $\S$\ref{sec:gapfilling-settings}, we train models across different $intersect_{=N}$ subsplits. We also subsample each split to be of equal sizes to keep the number of training examples consistent across the subsplits.

%



\subsubsection{Evaluation metric}
\label{sec:mlt-metric}
For the LT task, we propose a metric that rewards correctly translated ambiguous words and penalises words translated to the wrong sense. We name this metric \textbf{Ambiguous Lexical Index (ALI)}. An ALI score is awarded \emph{per English word} to measure how well the word has been correctly translated, 
and the score is averaged across all words. 
More specifically, for each 
English word, a score of $+1$ is awarded if the correct translation of the word is found in the output translation, a score of $-1$ is assigned if a known incorrect translation (from our dictionary) is found, and $0$ if none of the candidate words are found in the translation. The ALI scores for each 
English word is obtained by averaging the scores of individual instances, and an overall score is obtained by averaging across all 
English words. Thus, a per-word ALI score of $1$ indicates that the word is always translated correctly, $-1$ indicates that the word is always translated to the wrong sense, and $0$ means the word is never translated to any of the potential target words in our dictionary for the word. 

Being a per-word metric, another advantage of ALI is that you can choose only a subset of words to evaluate. For example, you may evaluate on only a subset of words that are highly ambiguous and more difficult. ALI will provide a measure of how difficult it is for a system to correctly translate this set of words to its correct sense in the target language on average, without overly biasing the evaluation towards words that appear more frequently in the test dataset. We will not test this in this paper, but will leave this as potential future work.

This metric is similar to the accuracy metric proposed by Lala \& Specia~\cite{LalaSpecia:2018}, where they evaluate the capabilities of MT systems at disambiguating words. However, they penalize equally translations with an incorrect sense (-1 in ALI) and translations that rephrase the text and do not contain any of the possible senses (0 in ALI). In addition, they only consider matches at token level, and do not consider matches at the lemma level. Finally, their metric is computed \emph{per instance}, while ALI is computed \emph{per English word}. 

An implementation of the ALI metric can be found in our evaluation script at\\ \url{https://github.com/josiahwang/multisubs-eval}.

\subsubsection{Results}
\label{sec:mlt-results}

\paragraph{Baseline models. } Like the fill-in-the-blank task, Table~\ref{tbl:mlt-baseline} reports the results of a simple $n$-gram with back-off baseline, trained on the full training set. The model will predict the translation given the target word and the $n-1$ words before the word in the source language, and will back-off as in the fill-in-the-blank model. The 1-gram model is equivalent to a Most Frequent Translation (MFT) baseline trained on the full dataset. We report the results for up to $n=5$, as the ALI scores are approximately the same beyond that. The ALI scores are generally quite high, and just by using one or two more context words in the source language (2-gram and 3-gram), we can already achieve reasonably high ALI scores.

\begin{table}[t]
    \caption{ALI scores on the lexical translation task for the $n$-gram with back-off baseline models trained on the full training set. The results are reported on the test subset. Note that 1-gram is equivalent to a Most Frequent Translation (MFT) baseline trained on the full dataset.
    }
    \label{tbl:mlt-baseline}
    \centering
    \begin{tabular}{l c c c c c}
        \toprule
        & 1-gram & 2-gram & 3-gram & 4-gram & 5-gram\\
        \midrule
        ES & 0.58 & 0.64 & 0.65 & 0.64 & 0.64\\
        PT & 0.54 & 0.64 & 0.69 & 0.69 & 0.69\\
        FR & 0.68 & 0.73 & 0.73 & 0.70 & 0.70\\
        DE & 0.50 & 0.56 & 0.59 & 0.59 & 0.59\\ 
        \bottomrule
    \end{tabular}
\end{table}

\paragraph{Neural-based models. } Table~\ref{tbl:mlt-ali} shows the ALI scores for our models, evaluated on the test set. We compare the model scores to a most frequent translation (MFT) baseline obtained from the respective $intersect_{=N}$ subsplits of the training data.
Interestingly, the MFT baselines actually performed better than the BRNN text-only model in general. The exception is for the $intersect_{=4}$ split, where the opposite is observed: the ALI scores for MFT drastically drop while the scores for the text-only model drastically improve. Further investigation is needed to ascertain the reason for the drastic change in scores. Our suspicion is that there are many English words in the test set with only a few test instances; ALI is computed per-word thus weights equally English words that occur frequently and those that occur once or twice in the test set. A variation in the predictions for these infrequent words, coupled with the small number of source words in the test set in general, might swing the scores drastically. For example, 31\% of the 143 source words in the Spanish test set have only 1-3 instances. 

For our models, we observe a general trend where as we go from $intersect_{=1}$ to $intersect_{=4}$ the ALI score over all models tend to consistently increase. The text-only models performed better than image-only models. Thus, as a lexical translation task, images do not seem to be as useful as text for translation. Indeed, like observations in multimodal machine translation research, textual cues play a stronger role in translation, as the space of possible lexical translations is already narrowed down by knowing the source word. There does not appear to be any significant improvement when adding image signals to text models. It still remains to be seen whether this is due to images not being useful or that the multimodal model is not effectively using images during translation.

It is also worth noting that our $n$-gram with back-off baselines trained on the full training set (Table~\ref{tbl:mlt-baseline}) achieved higher ALI scores than all our neural-based models (with the exception of the irregular score for the text and multimodal models trained on $intersect_{=4}$).

\begin{table}[t]
    \caption{ALI scores for the lexical translation task on the test set, comparing an MFT baseline, text-only, image-only, and multimodal models trained on different subsets of the data. 
    }
    \label{tbl:mlt-ali}
    \centering
    \resizebox{\linewidth}{!}{       
    \begin{tabular}{l c c c c c}
    \toprule
    & & \textbf{MFT} & \textbf{text} & \textbf{image} & \textbf{multimodal} \\
\midrule
\multirow{5}{*}{ES} & $intersect_{=1}$ & 0.53 & 0.42 & 0.19 & 0.43\\
 & $intersect_{=2}$ & 0.54 & 0.48 & 0.30 & 0.54\\
 & $intersect_{=3}$ & 0.64 & 0.55 & 0.33 & 0.51\\
 & $intersect_{=4}$ & 0.40 & 0.81 & 0.34 & 0.81\\
         \midrule
\multirow{5}{*}{PT} & $intersect_{=1}$ & 0.52 & 0.40 & 0.21 & 0.41\\
 & $intersect_{=2}$ & 0.55 & 0.47 & 0.30 & 0.45\\
 & $intersect_{=3}$ & 0.55 & 0.47 & 0.32 & 0.48\\
 & $intersect_{=4}$ & 0.36 & 0.79 & 0.37 & 0.80\\
\midrule
\multirow{5}{*}{FR} & $intersect_{=1}$ & 0.59 & 0.44 & 0.22 & 0.46\\
 & $intersect_{=2}$ & 0.66 & 0.50 & 0.30 & 0.54\\
 & $intersect_{=3}$ & 0.75 & 0.59 & 0.31 & 0.55\\
 & $intersect_{=4}$ & 0.31 & 0.81 & 0.33 & 0.81\\
 \midrule
\multirow{5}{*}{DE} & $intersect_{=1}$ & 0.35 & 0.41 & 0.27 & 0.43\\
 & $intersect_{=2}$ & 0.45 & 0.52 & 0.27 & 0.50\\
 & $intersect_{=3}$ & 0.58 & 0.54 & 0.34 & 0.53\\
 & $intersect_{=4}$ & 0.27 & 0.92 & 0.37 & 0.94\\
    \bottomrule
    \end{tabular}
    }
\end{table}


\section{Conclusions}
\label{sec:conclusion}

We introduced \multisubs, a large-scale multimodal and multilingual dataset aimed at facilitating research on grounding words to images in the context of their corresponding sentences. The dataset consists of a parallel corpus of subtitles in English and four other languages, and selected words are illustrated with one or more images in the context of the sentence. This provides a tighter local correspondence between text and images, allowing the learning of associations between text fragments and their corresponding images. The structure of the text is also less constrained than existing multilingual and multimodal datasets, making it more representative of multimodal grounding in real-world scenarios. 

Human evaluation in the form of a fill-in-the-blank game showed that the task is quite challenging, where humans failed to guess a missing word 42.41\% of the time, and could correctly guess only 21.89\% of instances without any images. We applied \multisubs on two tasks:  fill-in-the-blank and lexical translation, and compared automatic models that use and do not use images as contextual cues for both tasks. We plan to further develop \multisubs to annotate more phrases with images, and to improve the quality and quantity of images associated with the text fragments. \multisubs will benefit research on visual grounding of words especially in the context of free-form sentences, and is made publicly available under a Creative Commons licence on \url{https://doi.org/10.5281/zenodo.5034604}.

\begin{acknowledgements}
This work was supported by a Microsoft Azure Research Award for Josiah Wang. It was also supported by the MultiMT project (H2020 ERC Starting Grant No. 678017), and the MMVC project, via an Institutional Links grant, ID 352343575, under the Newton-Katip Celebi Fund partnership. The grant is funded by the UK Department of Business, Energy and Industrial Strategy (BEIS) and Scientific and Technological Research Council of Turkey (T{\"U}B{\.I}TAK) and delivered by the British Council. 
\end{acknowledgements}

\bibliographystyle{spmpsci}      
\bibliography{multisubs}   

\end{document}